\documentclass{article}
\usepackage{array}
\usepackage{subfigure}
\usepackage[T1]{fontenc}     
\usepackage{amsmath}
\usepackage{amssymb}
\usepackage{multirow}
\usepackage{graphicx}
\usepackage{url}
\usepackage{algorithm}
\usepackage{algorithmic}
\usepackage{setspace}
\usepackage{pdfpages}
\usepackage{epstopdf}
\usepackage{authblk}

\begin{document}

\title{A Deep Learning Approach with an Attention Mechanism for Automatic Sleep Stage Classification}


\author{Martin L\"angkvist, Amy Loutfi}

%
%


\affil{Center for Applied Autonomous Sensor Systems, \"Orebro University, Address\\
\"Orebro, Sweden\\
firstname.secondname@oru.se}

\date{}

\maketitle

\begin{abstract}
Automatic sleep staging is a challenging problem and state-of-the-art algorithms have not yet reached satisfactory performance to be used instead of manual scoring by a sleep technician. Much research has been done to find good feature representations that extract the useful information to correctly classify each epoch into the correct sleep stage. While many useful features have been discovered, the amount of features have grown to an extent that a feature reduction step is necessary in order to avoid the curse of dimensionality. One reason for the need of such a large feature set is that many features are good for discriminating only one of the sleep stages and are less informative during other stages. This paper explores how a second feature representation over a large set of pre-defined features can be learned using an auto-encoder with a selective attention for the current sleep stage in the training batch. This selective attention allows the model to learn feature representations that focuses on the more relevant inputs without having to perform any dimensionality reduction of the input data. The performance of the proposed algorithm is evaluated on a large data set of polysomnography (PSG) night recordings of patients with sleep-disordered breathing. The performance of the auto-encoder with selective attention is compared with a regular auto-encoder and previous works using a deep belief network (DBN).  
\end{abstract}


\section{Introduction}

The recent use of an attention mechanism for learning better internal representations has given promising results in a number of applications, such as speech recognition~\cite{ChorowskiBSCB15}, document comprehension~\cite{HermannKGEKSB15}, sentence summarization~\cite{RushCW15}, visual attention~\cite{288,xu2015show,MnihHGK14}, and machine translation~\cite{kalchbrenner2013recurrent,sutskever2014sequence,cho2014properties,bahdanau2014neural}. The attention mechanism allows the learning algorithm to focus on different parts of the input for each time frame. For sequence-to-sequence translations, the use of attention allows each predicted word in the output sentence to be dependent on all previous generated words and the selective attention on the whole input sentence instead of using a squashed fixed-length internal representation for the input sentence. For visual attention, the model adaptively shifts focus on parts of the image for generating the image description. The main advantage of using attention is that the model is reasonably capable of learning meaningful representations without reduced performance for very long input sentences or large input images. 

The task of automatic sleep stage classification is to classify 6-9 hours of multivariate time-series sleep data collected from a polysomnograph (PSG) into several categories of sleep stages. A common approach for solving this problem (see~\cite{267} for a review) is to choose a number of features and then perform feature selection to find an optimal subset of features for the current data set. Many of the used features try to capture the most relevant information for the current sleep stage and therefore mimic the standard Rechtschaffen and Kales (R\&K) system~\cite{7,8,4} that is manually used by sleep technicians.

In this work, we implement an attention mechanism on a sparse auto-encoder (SAE)~\cite{288} and a Recurrent Neural Network (RNN) with a long short-term memory cell (LSTM)~\cite{lstm1997} for learning context-relevant feature representations for the application of automatic sleep staging. 

This paper is organized as follows: The proposed model and the data is detailed in Section~\ref{sec:methods}. The experimental results are presented in Section~\ref{sec:results}. The related work to automatic sleep stage classification and attention mechanism is given in Section~\ref{sec:related}. Finally, the conclusion is given in Section~\ref{sec:conclusion}. The background to automatic sleep stage classification and feature extraction is given in~\ref{sec:appendix}.

\section{Material and Method} 
\label{sec:methods}

The modification weights the reconstruction error of the current inputs so that the cost for reconstruction error for suspected irrelevant inputs are reduced and the representational capacity is instead used on inputs that are suspected to be more informative for the current sleep stage. The modified auto-encoder, which we call selective attention auto-encoder, is used on a dataset of 25 PSG recordings and the classification result is compared with a standard auto-encoder and previous works using another representational learning algorithm, namely a deep belief network (DBN). The goal of this work is not to replicate the R\&K system or improve current state-of-the-art sleep stage classification but rather to explore the advantages of using a model with selective attention applied to automatic sleep staging. 

\subsection{Sparse Auto-Encoder}
An auto-encoder consists of an encoder and a decoder. The goal of the auto-encoder~\cite{29} is to reconstruct the input data via one or more layers of hidden units. The feed-forward activations in the encoder from the visible units $v_i$ to the hidden units $h_j$ is expressed as:
\begin{equation}
	h_j = \sigma_f\left(\sum_i W_{ji} v_i + b_j\right)
\end{equation}
where $\sigma_f$ is the activation function. In this work the sigmoid function is used which is defined as $\sigma_f(x) = \frac{1}{1+e^{-x}}$. In the decoder phase, the hidden layer is decoded back to reconstructions of the input layer. One pass of the decoder in layer $l$ is calculated as:
\begin{equation}
	\hat{v_i} = \sigma_g\left(\sum_j W_{ij} h_j + b_i\right)
\end{equation}
The activation function in the decoder can be the sigmoid function or the linear activation function $\sigma_g(x) = x$ if values in the input layer are not between $0$ and $1$. The cost function to be minimized for one training example is expressed as:
\begin{equation}
	L(v,\theta) = \frac{1}{2}\sum_i(v_i-\hat{v}_i)^2 + \frac{\lambda}{2}\sum_i\sum_j (W_{ji})^2 + \beta \sum_j KL(\rho||p_j) \label{eq:sae}
\end{equation}
where $\theta = \{W,b\}$ is the model parameters. The Kullback-Leibler (KL) divergence is defined as:
\begin{equation}
	KL(\rho||p_j) = \rho \log\frac{\rho}{p_j}+(1-\rho) \log \frac{1-\rho}{1-p_j} \label{eq:kldiv}
\end{equation}
where $p_j$ is the mean activation for hidden unit $j$ over all training examples in the current training mini-batch. The first term (reconstruction error term) ensures that the sum of the difference between the input units and the reconstructions of all input units over all training examples in the current training batch is small. The second and third regularization terms (weight decay term and sparsity penalty term) prevents the trivial learning of a 1-to-1 mapping of the input and comes with one or more hyperparameters ($\lambda$, $\beta$, $\rho$). The hyperparameters are set with a random grid search~\cite{61}.

\subsection{Sparse auto-encoder for Time-Series}
\begin{figure*}[ht]
 \centering
    \subfigure[Auto-encoder]{\includegraphics[width=0.45\textwidth]{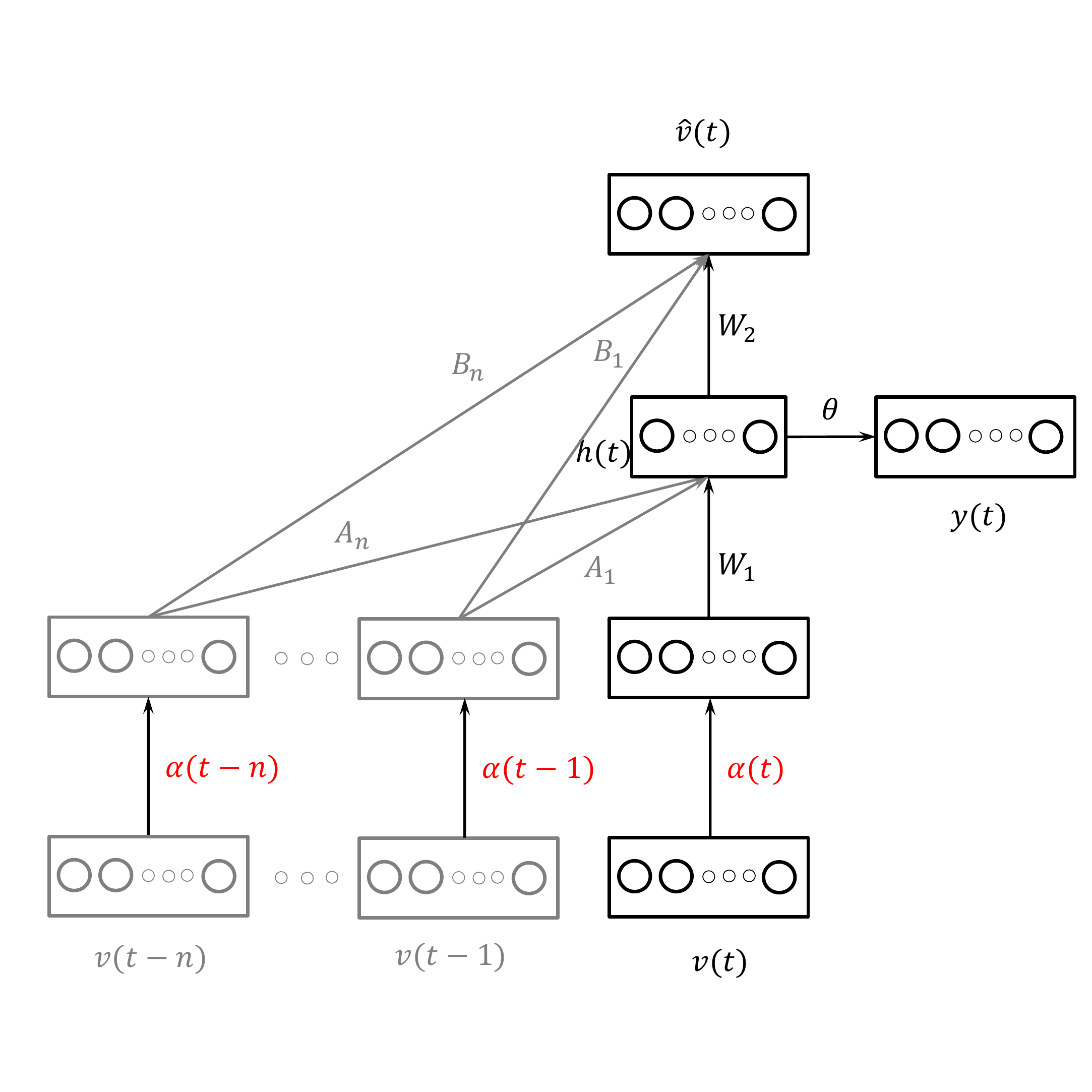} \label{fig:modelsAEbefore}}
    \subfigure[Recurrent Neural Network]{\includegraphics[width=0.45\textwidth]{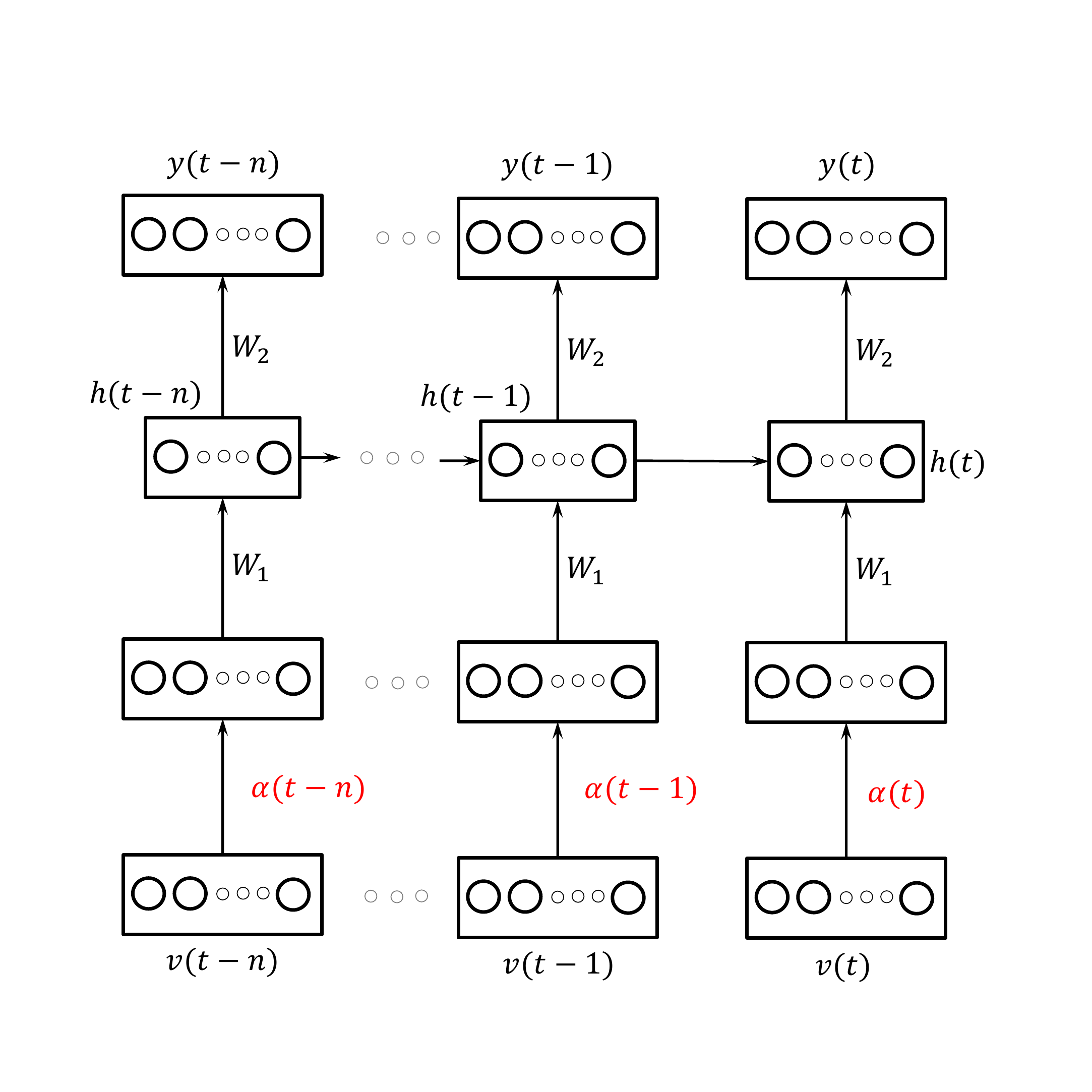} \label{fig:modelsRNNbig}}
    \caption{(a) Unsupervised feature learning for a single-layer auto-encoder for structured data with attention mechanism. (b) Supervised learning for a Recurrent neural network with attention mechanism.}
  \label{fig:modelsafter}
\end{figure*}

The auto-encoder can be extended that shares a similar structure to a conditional Restricted Boltzmann Machine~\cite{37} in order to capture temporal structure in sequential data, see Figure~\ref{fig:modelsAEbefore}.

In a temporal auto-encoder, the hidden units depend on the visible units of the current timeframe as well as visible units of previous timeframes. The hidden layer at time $t$ is calculated as:
\begin{equation}
	h_j = \sigma_f\left(\sum_{k=1}^n\sum_i A^k_{ji} v_i(t-k) + \sum_i W_{ji} v_i + b_j\right)
\end{equation}
where $n$ is the model order and $A^n$ is the weight matrix between the hidden layer and visible units at time frame $t-n$. The reconstruction layer is calculated as:
\begin{equation}
	\hat{v}_i = \sigma_g\left( \sum_{k=1}^n\sum_j B^k_{ji} v_i(t-k) + \sum_j W_{ij} h_j + b_i\right)
\end{equation}
where $B^n$ is the weight matrix between visible units at time frame $t-n$ and the reconstruction of the visible layer at the current time frame $t$. There is no reconstruction of past visible layers. The past visible layers act as an extra bias term, much similar to a conditional RBM~\cite{37}.

\subsection{Selective Attention Sparse Auto-Encoder for Time-Series}
\label{sec:sasae}

With the formulation of the reconstruction error term in Eq.~\ref{eq:sae}, the learning algorithm attempts to reconstruct all input units equally. This paper uses a method that introduces selective attention by reducing the reconstruction error cost for a selected number of inputs. The selective attention is different depending on the category that the current training data belongs to. This is implemented by introducing the weighting vector, $\alpha_i^k$, which indicates the probability that unit $i$ should be reconstructed if the input belongs to class $k$. If $\alpha^k_i = 1$ for $\forall i,k$ the model generalizes to a regular sparse auto-encoder. The small change to the first term in Eq.~\ref{eq:sae} is:
\begin{equation}
	L(v, \theta, \Lambda, k) = \frac{1}{2}\sum_i(v_i-\hat{v_i})^2 \cdot \Lambda^k_i  + \frac{\lambda}{2}\sum_i\sum_j (W_{ji})^2 + \beta \sum_j KL(\rho||p_j)
\end{equation}
where $\Lambda^k_i$ is set to 1 with probability $\alpha_i^k$ and set to 0 otherwise.
\begin{align}
P(\Lambda^k_i=1|\mathbf{\alpha})&=\alpha_i^k \\
P(\Lambda^k_i=0|\mathbf{\alpha})&=1-\alpha_i^k
\end{align}
This work explores two methods for setting the values of $\alpha_i^k$. The first method is when the values are fixed and can be set with a feature selection algorithm. In this work the t-test algorithm is used where the test statistic is calculated as:
\begin{equation}
t = \frac{\mu_1-\mu_2}{\sqrt{\frac{\sigma_1^2}{n_1}+\frac{\sigma_2^2}{n_2}}}
\end{equation}
where $\mu_i$, $\sigma_i$, and $n_i$ is the mean, standard deviation and number of examples of a variable that belongs to class $i$. For more than two classes the test statistic is calculated one-vs-all for each class.

The second method for setting the values of $\alpha_i^k$ is to learn them together with the model parameters. This is done by introducing an weighting penalty term, $f(\alpha_i)$, which is a function that adds a cost for deviating the values of $\alpha_i^k$ from the starting values. In this work we use the same Kullback-Leibler divergence from Eq.~\ref{eq:kldiv} with $\rho=1$. The advantage of the KL-divergence penalty is that it has a asymptote at 0 which will keep the values above 0. With the added penalty term comes a new hyperparameter, $\gamma$. 

\subsection{Experimental Data set}
The data set that is used in this work has kindly been provided by St. Vincent's University Hospital and University College Dublin and is freely available for download at PhysioNet~\cite{167}. The data set consists of 25 acquisitions from subjects with suspected sleep-disordered breathing. Each acquisition consists of 2 EEG channels (C3-A2 and C4-A1), 2 EOG channels, and 1 EMG channel. Sample rate is 128 Hz for EEG and 64 Hz for EOG and EMG. Scoring was manually performed by one sleep expert.

The training, validation, and test sets are created by randomly split the 25 acquisitions into sizes of $60\%/20\%/20\%$, respectively. A 5-fold cross-validation is performed with different random splits.

\section{Results and Discussion} 
\label{sec:results}
The model that is used in this work is a 1-layer auto-encoder. The size of the model is set to 500 hidden units and is trained on the training set until the output from the cost function on the validation set has not decreased for 10 epochs. One training epoch consists of going through all mini-batches of the training set. Each mini-batch consists of 30 randomly selected 30-seconds segments in order to have data from each class in each mini-batch. Stochastic gradient descent with momentum ($0.9$) and decaying learning rate ($0.01$) is used as optimization method. 

The test set is created by randomly drawing 5 of the 25 full-night PSG recordings. Different test sets are used to perform 5-fold cross-validation. The validation set is created by randomly drawing 5 of the remaining 20 acquisitions from the training set. 

The hyperparameters are set with random grid search~\cite{61}. For each simulation, each hyperparameter is randomly set from a list of possible choices and after a number of simulation the combination that gave the highest classification accuracy on the validation set is selected. The hyper parameters were chosen from $\lambda = \{10^{-2}, 10^{-3}, 10^{-4}, 10^{-5}\}$, $\beta = \{3, 0.3, 0.03\}$, and learning rate $\eta = \{10^{-3}, 10^{-4}, 10^{-5}\}$. The sparsity parameter $\rho$ is set to $0.05$ meaning that each hidden unit should aim to be "active" $5\%$ of the time. 

The values of the weighting vector $\alpha_i^k$ are set in three different ways: (standard) all values are are set to 1 for all $i$ and $k$ and are not updated (this is the same as a standard auto-encoder); (fixed) the values are set using the absolute and normalized raw scores from the t-test feature selection algorithm and are not updated; (adaptive) all values are initially set to $1$ and then updated together with the model parameters during learning according to Section~\ref{sec:sasae}. 

Training is first performed with unsupervised pre-training and one of the three choices of weighting vector. The decoding part of the auto-encoder is then removed and a layer of softmax units is attached on the hidden layer in order to perform supervised finetuning and classification. The weighting vector has no effect on the learning during the supervised finetuning phase. The trained model is then used to perform feed-forward classification on the test set. 

\begin{figure}[ht]
 \centering
    \subfigure[]{\includegraphics[width=0.45\textwidth]{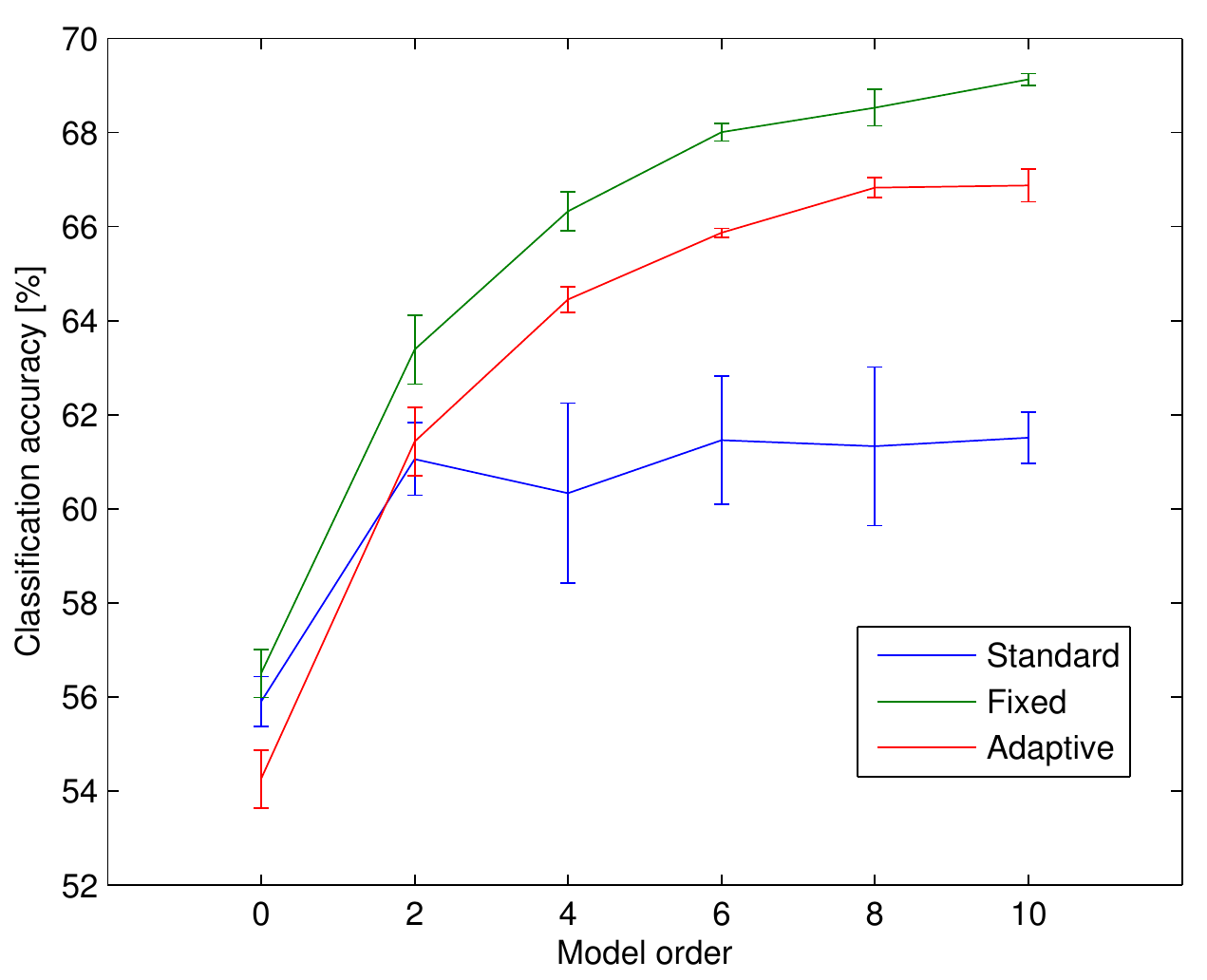} \label{fig:Nresultsa}} \quad
    \subfigure[]{\includegraphics[width=0.45\textwidth]{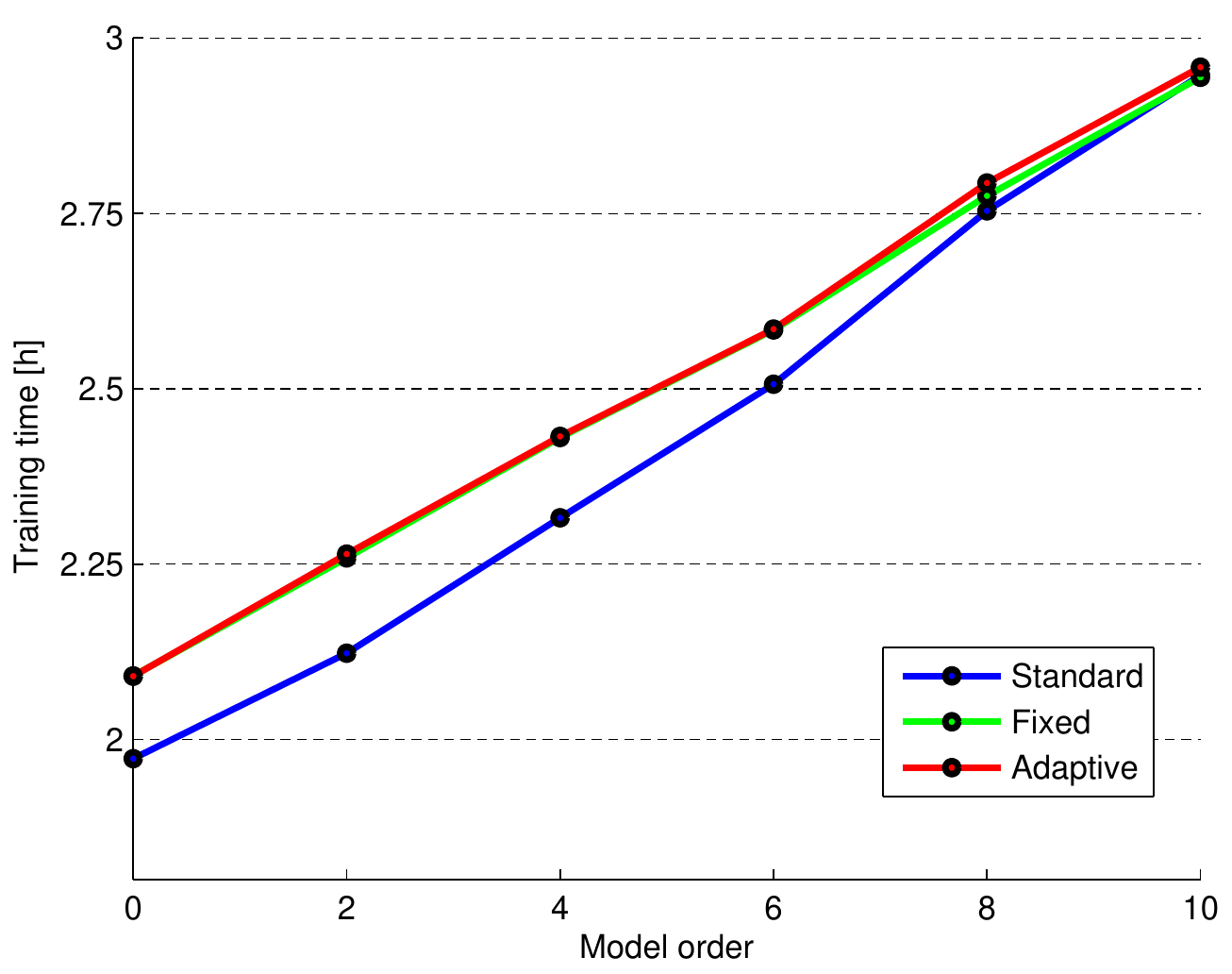} \label{fig:Nresultsb}}
    \caption{(a) Classification accuracy (mean and standard deviation from 5-fold cross-validation) and (b) training time for models with varying model order.}
  \label{fig:Nresults}
\end{figure}

Figure~\ref{fig:Nresults} shows the classification results when using the three types of standard, fixed, and adaptive weighting vector for different model orders. The model order is the number of time steps $n$ from Figure~\ref{fig:modelsafter}. It can be seen that the fixed and adaptive weighting achieves a higher classification accuracy on the test set than a standard auto-encoder, except for when the model order is zero. The fixed weighting vector gave better classification accuracies than the adaptive weighting vector but the adaptive method outperforms the standard auto-encoder for the higher model order. The hyperparameters where not individually optimized for each method and model order because of the long training time. The training time can be seen in Figure~\ref{fig:Nresultsb} and shows that for low model order the standard auto-encoder is faster to train but the difference decreases as the model order is increased. The reason for this is that the time for updating the weighting vector is constant and is not affected by the model order. 

\begin{figure}[ht]
 \centering
    \subfigure[]{\includegraphics[width=0.3\textwidth]{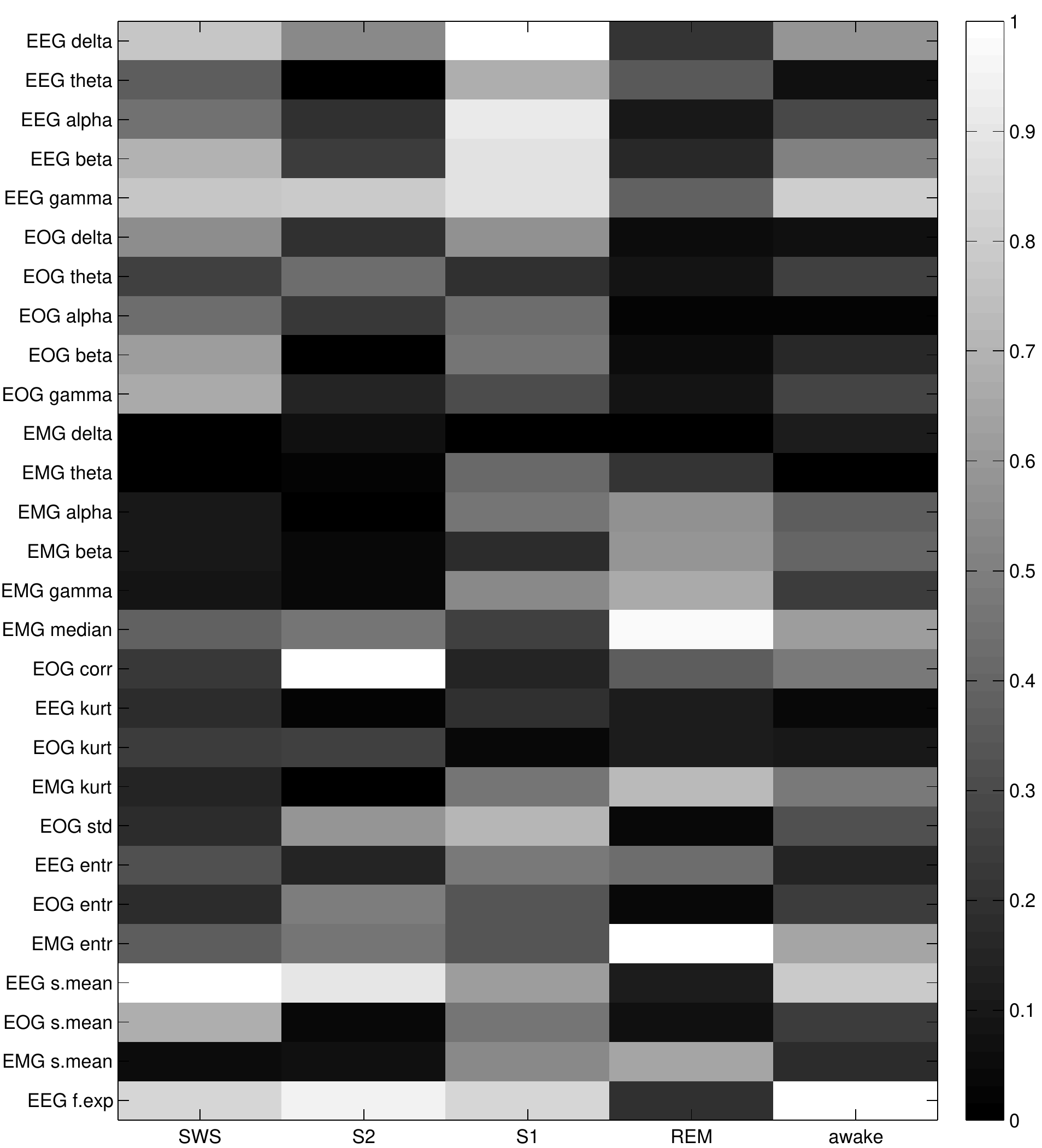} \label{fig:alphavaluesa}} \quad
    \subfigure[]{\includegraphics[width=0.3\textwidth]{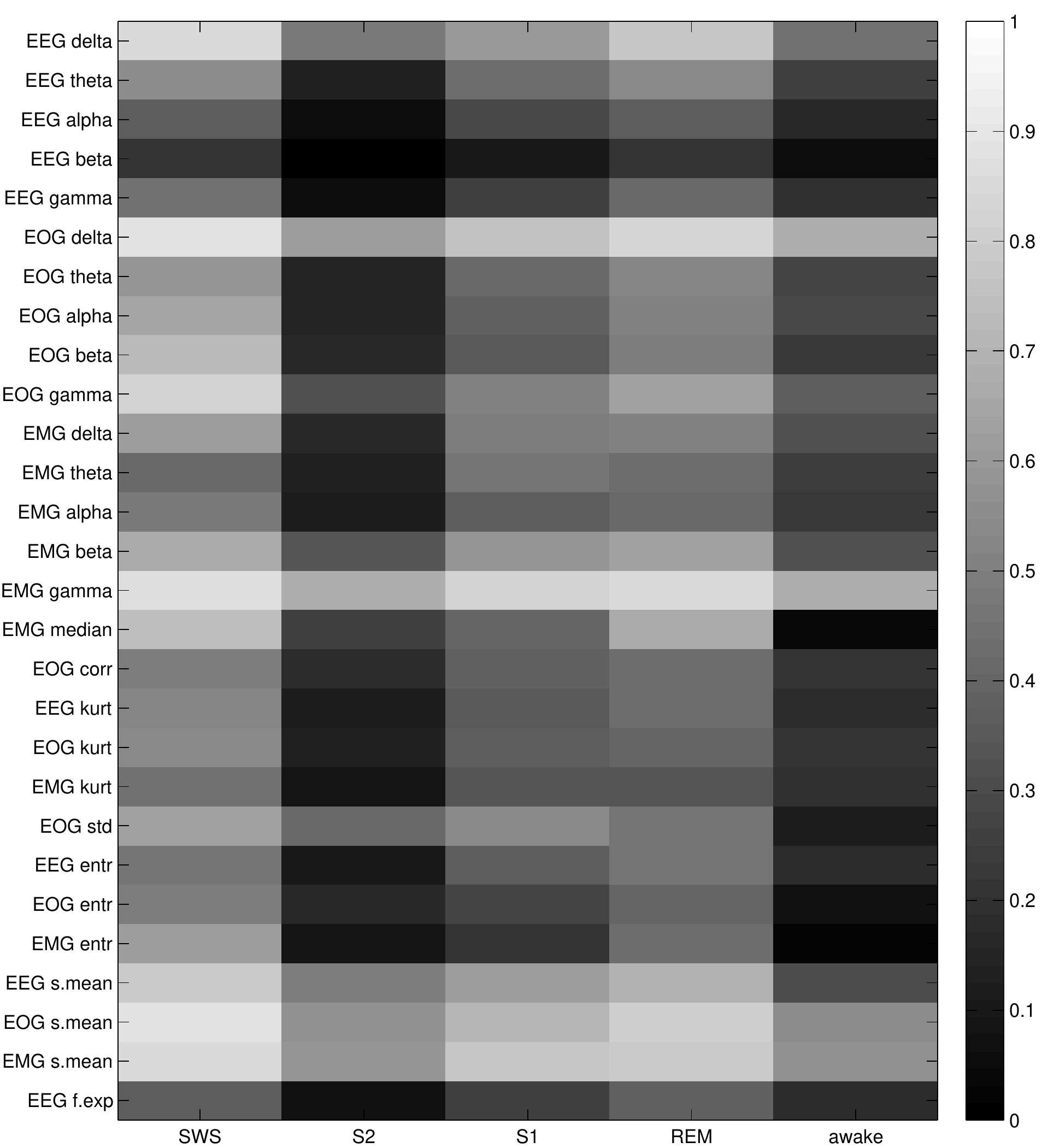} \label{fig:alphavaluesb}}
		\subfigure[]{\includegraphics[width=0.3\textwidth]{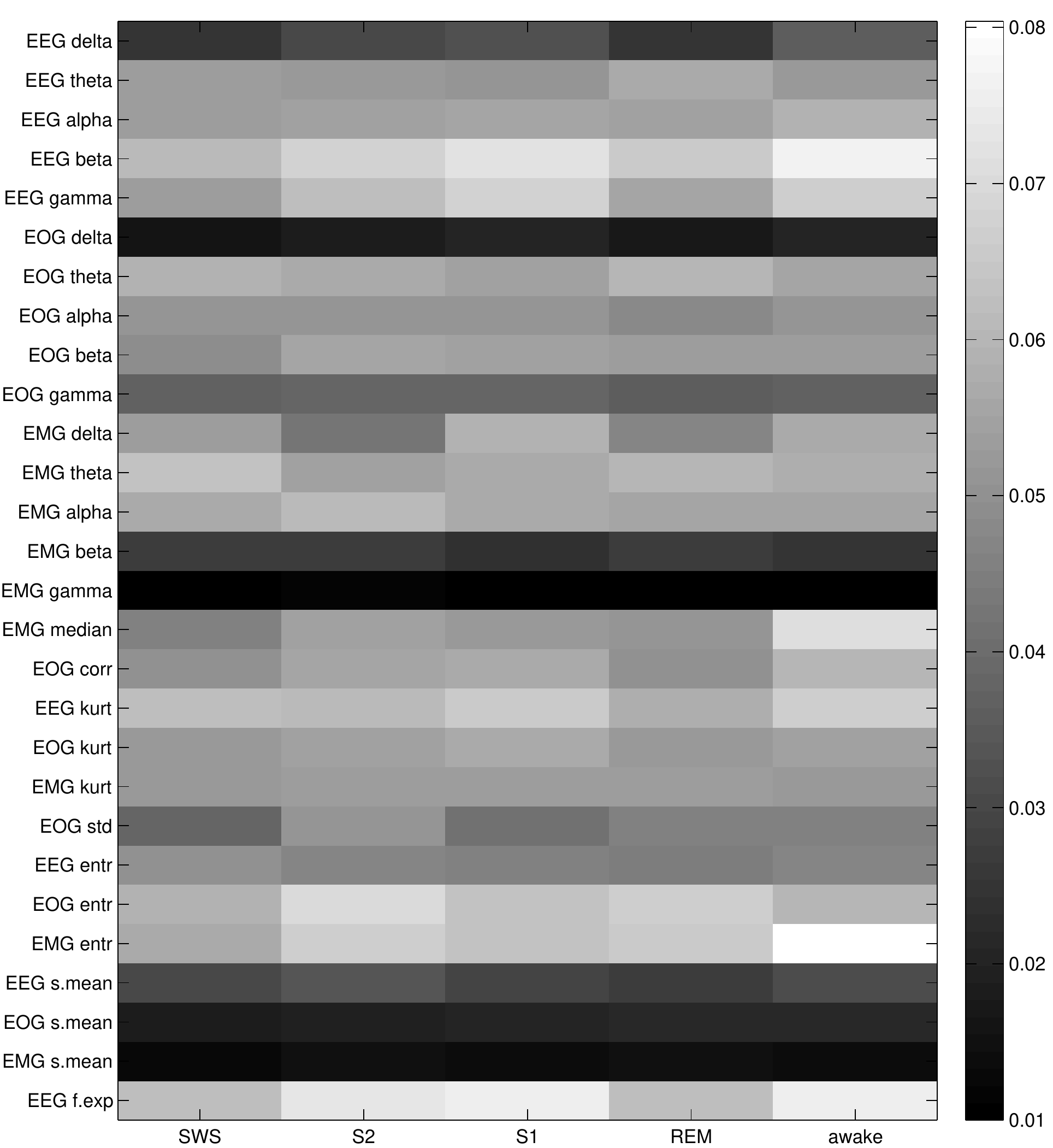} \label{fig:RecErrorStandard}}
    \caption{(a) Values of fixed weighting vector. The method can totally ignore one feature for one sleep stage and fully reconstruct it in another sleep stage. (b) Values of adaptive weighting vector. The method focuses more on some features and sleep stages than others. The difference of the values across sleep stages is not as high as for the fixed method. (c) The average reconstruction error from a standard auto-encoder on the validation set grouped by each sleep stage. Some features are harder to reconstruct at some sleep stages than others. A higher average reconstruction error for one feature at one sleep stage gives a lower probability that that feature should be reconstructed at that sleep stage when the adaptive approach is used.}
  \label{fig:alphavalues}
\end{figure}

Figure~\ref{fig:alphavalues} shows values of the weighting vector when the weights are either fixed or adaptive. The fixed values (Figure~\ref{fig:alphavaluesa}) were set with the t-test feature selection algorithm on the inputs and labels from the training set. Each feature is assigned both high values and low values depending on the current sleep stage. For example, the priority to reconstruct the features that describe the frequency of the EEG is higher for stage 1 (where the amount of alpha-waves is decreasing) than in REM-sleep (which has mixed frequency waves). The values for the EMG features are higher at the REM stage, in particular the median and entropy, than during other stages of sleep. The EEG and EOG features all have low values for determining REM-sleep. The feature that measures the correlation between the EOG channels have highest values for stage 2 (where the slow eye-movements have disappeared). Some features have a low value for every sleep stage, e.g., EMG delta, EEG kurtosis, and EOG kurtosis. 

The final adaptive weighting vector has a different look, see Figure~\ref{fig:alphavaluesb}. Here, the labels are not used to set the values of the weighting vector but instead the values are learned together with the model parameters during learning. It can be seen that there is a structure of vertical and horizontal lines where each feature (or sleep stage) generally has a high or low value across each sleep stage (or feature). The values for stage 1 and awake are generally lower than the other stages and some features (EEG and EOG delta, EOG and EMG gamma, all three spectral means) have higher values.

The adaptive weighting vector is determined by how easy the inputs can be reconstructed during the different sleep stages. A lower probability to reconstruct inputs is given to inputs that have a high reconstruction error. Figure~\ref{fig:RecErrorStandard} shows the average reconstruction error on the validation set for each sleep stage with a standard auto-encoder. Some features (EOG delta, EMG gamma, EOG and EMG spectral mean) are easy to reconstruct regardless of the sleep stage, while other features are easier to reconstruct for some stages and harder to reconstruct at others, for example EMG entropy and median is harder to reconstruct during awake stage. Since the standard auto-encoder is trained to reconstruct all inputs equally for all sleep stages, the difference in the average reconstruction error across the sleep stages indicates that some features are unpredictable, i.e., they behave like noise. 

\begin{figure}[!t]
\centering
\includegraphics[width=0.8\textwidth]{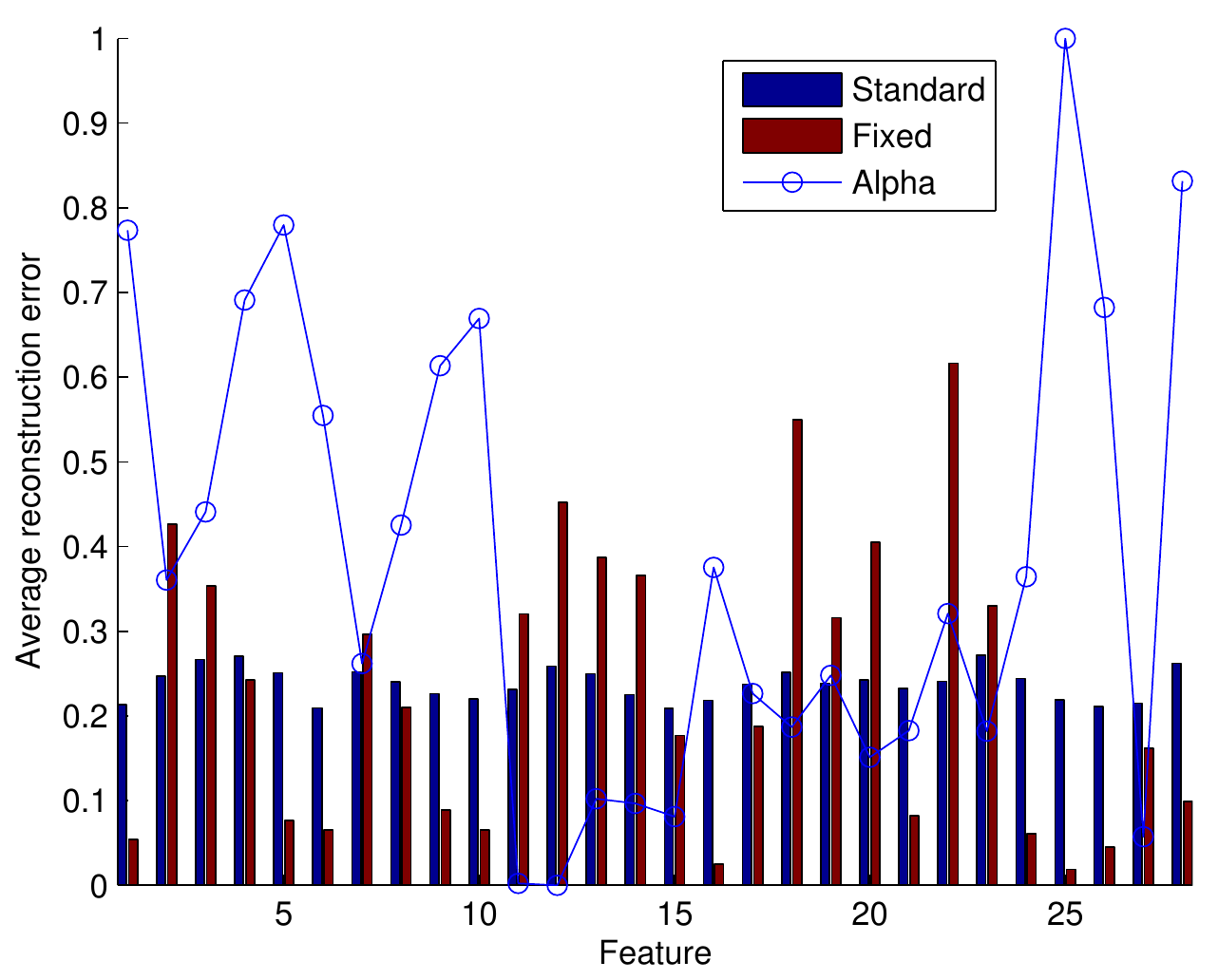} \label{fig:RecError}
\caption{Average reconstruction error for each input unit for standard and fixed weighting vector for slow-wave sleep stage. The error is generally increased for low values of $\alpha$ and decreased for higher values of $\alpha$.}
\end{figure}

Figure~\ref{fig:RecError} shows the average reconstruction error for standard and fixed weighting vector for the slow-wave sleep stage. The average reconstruction error for each input unit is around $0.2$ when a standard auto-encoder is used, i.e, each input is treated equal. With a fixed weighting vector the reconstruction error is lower for the inputs (features) that have a higher value of $\alpha$, and vice versa. 

For comparison, Table~\ref{table:accHMM} shows the average classification accuracy over a 5-fold cross validation using the three choices of weighting vector. For temporal smoothing, the predicted class of each 1-second segment from the training set is used to train a Hidden Markov Model which is then used to smooth out the sleep transitions from the classified 1-second segments from the test set. A selective attention auto-encoder using either the adaptive or fixed method achieved a higher mean classification accuracy compared to a standard auto-encoder. The fixed method achieves a higher classification mean than a 2-layered deep belief net (DBN) with 200 hidden units in each layer. Due to the low reliability of the score given that the used data set was only scored by one human expert and that scoring of sleep data generally has an inter-rater reliability of $80\%$, the classification accuracy may seem low compared to other automatic sleep staging methods. While those works normally carefully select unambiguous epochs for test set, the aim of this work is not to beat current automatic sleep staging methods on benchmark data sets but to explore the feasibility of applying the method of selective attention to a multivariate time-series problem.
\begin{table*}[!t]
\caption{Classification accuracy (mean$\pm$std[\%]) with and without temporal smoothing for a deep belief net (DBN) and sparse auto-encoder (SAE) with different methods of setting the weighting vector.}
\label{table:accHMM}
\centering
\begin{tabular}{|l|c|c|} \hline
Method & Without temporal smoothing & With temporal smoothing\\ \hline
DBN~\cite{65} 						& - 							& $72.2\pm9.7$ \\
SAE (standard $\alpha$) 	& $66.9\pm4.94$ 	& $71.9\pm5.2$ \\
SAE (adaptive $\alpha$) 	& $70.3\pm6.55$ 	& $76.5\pm6.5$ \\
SAE (fixed $\alpha$) 			& $71.0\pm5.93$   & \textbf{77.7$\pm$6.9} \\ \hline
\end{tabular}
\end{table*}

The result of the sparse auto-encoder with fixed weighting vector can be further analyzed by examining the confusion matrix, see Table~\ref{table:confmat}. Here it can be seen that stage 1 (S1) is the hardest class to classify and the biggest confusion is S1 being classified as S2 or awake. Similarly, there is a confusion about awake data being classified as S1. 
\begin{table}[!t]
		\caption{Confusion matrix for SAE (fixed $\alpha$).}
\label{table:confmat}
\centering
\begin{tabular}{cc|c c c c c}
     & & \multicolumn{5}{c}{Classified} \\
     & \% & awake & S1 & S2 & SWS & REM \\ \cline{2-7}
     \multirow{5}{*}{\rotatebox{90}{Expert}} 
  & awake & 73.5 &  18.4  &  5.8  &  0.6  &  1.7 \\
  & S1    &  9.6 &  60.8  & 20.3  &  0.8  &  8.5 \\
  & S2    &  0.6 &   7.0  & 81.9  &  6.0  &  4.6 \\
  & SWS   &    0 &      0 &   9.1 &  90.9 &    0 \\
  & REM   &  2.6 &   1.9  &  7.0  &  0    & 88.5 \\
     \end{tabular}
\end{table}

\section{Related Work} 
\label{sec:related}

The authors of~\cite{277} developed an automatic 6-stage (slow wave sleep was divided into stage 3 and 4) classification system that achieved $79.6\%$ accuracy on all 30-second epochs that where decided as scoreable. The data consisted of one EEG channel, two EOG channels, and one EMG channel from the Siesta polysomnographic database~\cite{257}. Training and validation set was equally split from 572 recordings of both healthy patients and patients with sleep disorders of adults between age 20 and 95. The sleep stager extracts features that follows the decision rules for visual scoring by looking for known markers such as sleep spindles, delta waves, slow eye-movements (SEMs) and rapid eye-movements (REMs). The automatic stager also uses prior knowledge such as a raw data and feature quality check, prior probabilities of stage changes, movement detection, position of the epoch within the NREM/REM sleep cycle, a rule-based smoothing procedure for transition to and from REM stage, and a comparison to other subjects within the same age and sex group as the current test subject. While this method and similar approaches~\cite{6,9,10,32,258,263,264,266,268,276} has been done, applying such techniques has not been standardized for use in a clinical setting and therefore one can argue that no set of universally applicable set of features has yet been found~\cite{267,260}. 

New features that are not obvious from the rule-based definitions of the sleep stages have also been discovered, e.g., ratios between frequency bands~\cite{266} and fractal exponent~\cite{263}. An alternative approach to the method that adheres to the R\&K system or researching for new hand-made features is to use unsupervised learning. This has previously been done for automatic sleep staging with for example power spectrum analysis~\cite{269,270} and metric learning with a Large Margin Nearest Neighbor on a k-NN classifier~\cite{271}. The latter approach achieved $94.4\%$ classification accuracy (when stage 1 and REM was grouped together) on the Sleep-EDF data set~\cite{167}.

Another promising method is to use representational learning algorithms~\cite{23,60,76,40} that automatically constructs the features from the input data. The advantage of these methods is that they are capable of modeling high-dimensional complex data by constructing it's own internal representation from unlabeled data. They have already been applied to various multivariate time-series problems (see~\cite{261} for a review). Some examples include speech recognition~\cite{149}, music recognition~\cite{170}, motion capture data~\cite{37}, gas identification with an electronic nose~\cite{278}, emotion classification~\cite{272,273,275},  rhythm perception~\cite{279}, and Brain Computer Interface (BCI) applications~\cite{274}. Representational learning algorithms also have been used for modeling PSG recordings. The work in~\cite{35} trained a deep belief network (DBN) on EEG signals for anomaly detection and the work in~\cite{65} uses a similar approach for the task of sleep stage classification using both raw data and pre-defined features as input. 

One challenge with unsupervised learning algorithms is that for one training example all input data in that training example is treated equally. In many multivariate time-series problems, including PSG recordings, there may be signals that are redundant or less informative than others. The traditional approach is to identify such signals and remove them. However, in many cases, there is no signal that is useful for all categories and often a signal is not totally redundant and may instead be the deciding factor for discriminating between two categories. For example, the EEG channels show a similar appearance in both stage 1 and REM stage but it is mostly the amplitude of the EMG that is the deciding factor. But the role of the EMG amplitude is not as crucial for discriminating other stages of sleep for example stage 1 and 2~\cite{265}. This means that some signals, and features based on those signals, should have less impact on the sleep stage decision depending on the current sleep stage. 

\section{Conclusion}
\label{sec:conclusion}
In this work we have shown that a per-category selective feature attention improves feature learning for the task of classifying sleep stages. Two different methods for setting the selective attention has been explored: a static approach that fixed the weighting vector with a supervised feature selection algorithm on the input data, and an adaptive approach that in an unsupervised fashion learned the area for selective attention during learning. The fixed approach outperformed the adaptive approach since it uses the knowledge of the correct labels but the adaptive approach achieved better classification than a standard auto-encoder. The main advantage of the proposed method, regardless of choice of method for selecting the selective attention, is that there is no need to perform feature selection, which is usually the most focused area in automatic sleep stage classification.

An interesting direction for future work is to explore the use a larger pool of features or using the raw signals as input instead of features. 

\section*{Acknowledgment}
The authors are grateful to Professor Walter T McNicholas of St. Vincents University Hospital, Ireland and Professor Conor Heneghan of University College Dublin, Ireland, for providing the sleep training data for this study. We would also like to thank senior physician Lena Leissner and sleep technician Meeri Sandelin at the sleep unit of the neuro clinic at \"Orebro University Hospital for their continuous support and expertise. 

\appendix
\section{Feature extraction for sleep stage classification}
\label{sec:appendix}

The amount of quality sleep has a decisive influence on health, behavior, mood~\cite{257}, as well as concentration, decision-making, and learning~\cite{258}. Diagnosing sleep disturbances requires a number of psychophysiological parameters that can be obtained by long-term activity monitoring, maintaining a sleep diary, performing psychometric tests, or analysis of polysomnographic (PSG) recordings by a sleep technician. A PSG recording consists of channels of electroencephalography (EEG), electrooculogram (EOG), and electromyogram (EMG), together with physiological parameters, such as oxygen saturation of arterial blood, electrocardiography (ECG), excursion of chest and abdomen, nasal airflow, and limb movements. A hypnogram is created by manually label each epoch of 20 or 30 seconds into one of the five sleep stages (wake (W), stage 1 (S1), stage 2 (S2), slow wave sleep (SWS) and rapid eye-movement (REM) sleep) defined by Rechtschaffen and Kales (R\&K)~\cite{7,8,4}. A summary of the characteristics for the different sleep stages can be seen in Table~\ref{table:sleepdefinition}.
\begin{table*}[!t]
\caption{Summary of definition for each sleep stage.}
\label{table:sleepdefinition}
\centering
\begin{tabular}{|l|p{3.0cm}|l|p{4.9cm}|} \hline
Stage & EEG & EMG & EOG \\ \hline
W 		& low amplitude, >50\% alpha waves 							& high amplitude			& slow eye-movements (SEMs), reading eye-movements, or blinking \\ \hline
S1 		& low amplitude, <50\% alpha waves  					  & medium amplitude		& possible SEMs \\ \hline
S2 		& K-complexes (KC) and sleep spindles (SS) 			& medium amplitude		& no SEMs \\ \hline
SWS 	& >20\% delta waves															& low amplitude 			& typically no eye-movements \\ \hline
REM 	& no KC or SS, low amplitude, mixed frequencies & lowest amplitude 		& rapid eye-movments (REM) \\ \hline
\end{tabular}
\end{table*}

The signals and feature extraction follows a previous work on the same data set from the same authors~\cite{65}. All signals are pre-processed by notch filtering at 50 Hz in order to cancel out power line disturbances and down sampled to 64 Hz after being pre-filtered with a band-pass filter of 0.3 to 32 Hz for EEG and EOG, and 10 to 32 Hz for EMG.

A total of 28 features are extracted from each 1-second segment with zero overlap in the 4-channel (1 EEG, 2 EOGs, 1 EMG) PSG recording. The used features are relative power for five frequency bands (delta ($0.5-4 Hz$), theta ($4-8 Hz$), alpha ($8-13 Hz$), beta ($13-20 Hz$), and gamma ($20-32 Hz$)) of all signals, median of EMG, standard deviation of one EOG, correlation coefficient between both EOGs, entropy, kurtosis, and spectral mean of all signals, and fractal exponent~\cite{17,33} of EEG. All features are first transformed with a non-linear transformation~\cite{18} and then normalized with z-score~\cite{262}. A summary of the used features can be seen in Table~\ref{table:features}.
\begin{table*}[!t]
\caption{Calculation of used feature $y$ from input signal $x$. The five frequency bands (delta, theta, alpha, beta, and gamma) are noted with $f_1$-$f_5$. The mean and standard deviation of the input signal is noted with $\mu$ and $\sigma$, respectively.}
\label{table:features}
\centering
\begin{tabular}{|p{3.2cm}|c|c|} \hline
Feature, $y$ & Calculation & Channels, $x$ \\ \hline
Relative power 	& $\frac{P(x,f_i)}{\sum_{f={f_1}}^{f_5} P(x,f)}$ & EEG, EOG1, EMG\\ \hline
Median & $median(|x|)$ & EMG \\ \hline
Correlation & $\frac{E\left[(x_1 - \mu_{x_1})(x_2 - \mu_{x_2})\right]}{\sigma_{x_{1}}\sigma_{x_{2}}}$ & EOG1, EOG2 \\ \hline
Entropy & $-\sum_i x_i^2 \ln x_i^2$ & EEG, EOG1, EMG \\ \hline
Kurtosis & $\frac{E[x-\mu]^4}{\sigma^4}$ & EEG, EOG1, EMG \\ \hline
Spectral mean 	& $\frac{1}{5}\sum_{f=f_1}^{f_5} P_{rel}(x,f) \cdot |f|$ & EEG \\ \hline
\end{tabular}
\end{table*}

\begin{figure*}[!t]
\centering
\includegraphics[width=0.9\textwidth]{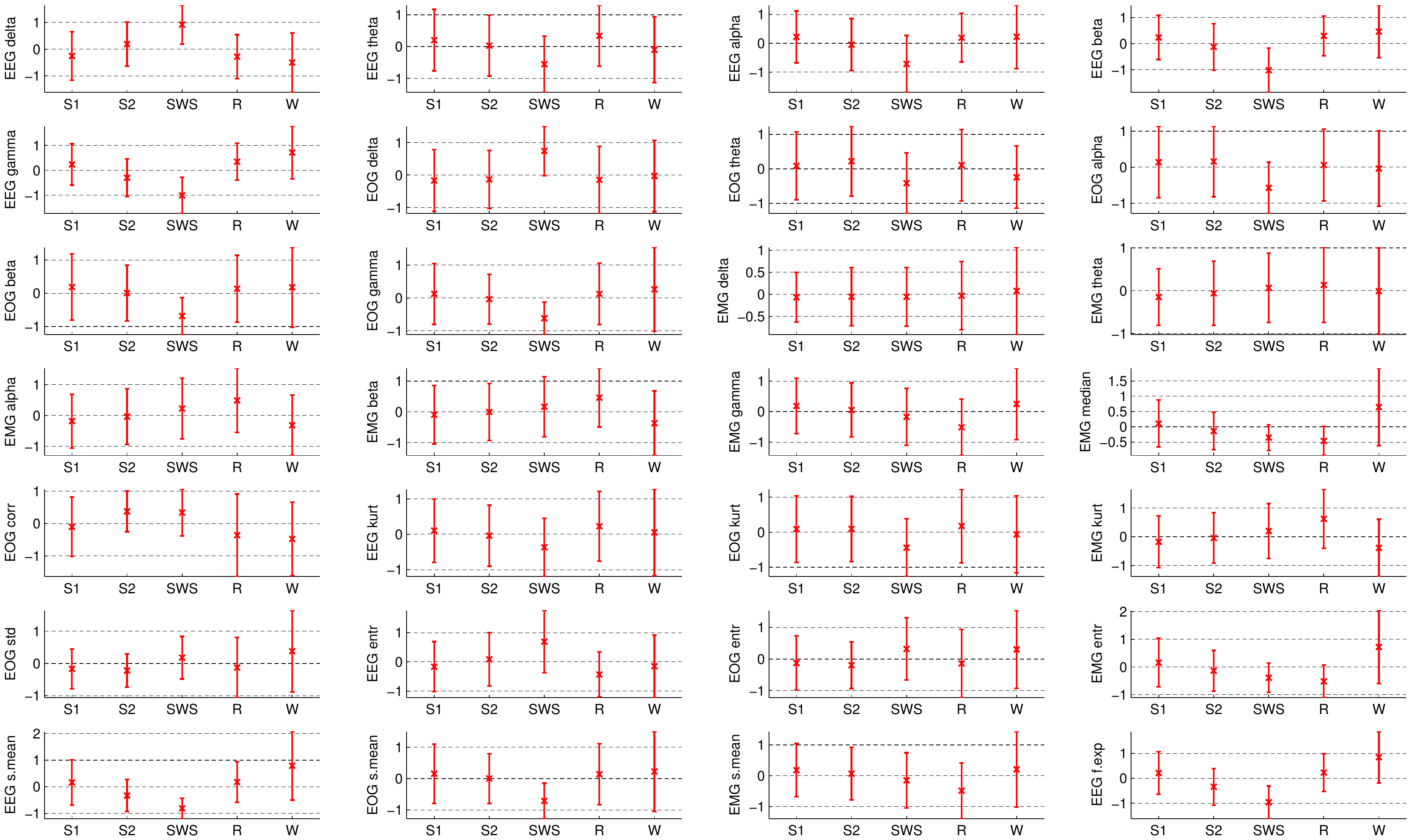} \label{fig:errorbarfeat}
\caption{Error bar (mean and standard deviation) for all 28 features grouped by sleep stage. A good feature for discriminating between different sleep stages has non-overlapping values in those sleep stages, for example EEG gamma is a good feature for discriminating between SWS and REM/awake. A poor feature typically has a high standard deviation at all sleep stages and equal mean, for example EMG delta and EMG theta.}
\end{figure*}

Figure~\ref{fig:errorbarfeat} shows the mean and standard deviation of the normalized values of all 28 features for each of the 5 categories. It can be seen that some features alone are good indicators for certain sleep stages. For example, spectral mean of EEG is a good indicator for slow-wave sleep, fractal exponent of EEG is a good discriminator between wake and slow-wave sleep, and median and entropy of EMG are good indicators for REM-sleep. 

\bibliographystyle{plain}
\bibliography{mybibfile}

\begin{thebibliography}{10}

\bibitem{277}
Peter Anderer, Georg Gruber, Silvia Parapatics, Michael Woertz, Tatiana
  Miazhynskaia, Gerhard Kl{\"o}sch, Bernd Saletu, Josef Zeitlhofer, Manuel~J
  Barbanoj, Heidi Danker-Hopfe, et~al.
\newblock An e-health solution for automatic sleep classification according to
  rechtschaffen and kales: validation study of the somnolyzer 24$\times$ 7
  utilizing the siesta database.
\newblock {\em Neuropsychobiology}, 51(3):115--133, 2005.

\bibitem{bahdanau2014neural}
Dzmitry Bahdanau, Kyunghyun Cho, and Yoshua Bengio.
\newblock Neural machine translation by jointly learning to align and
  translate.
\newblock {\em arXiv preprint arXiv:1409.0473}, 2014.

\bibitem{40}
Yoshua Bengio.
\newblock Learning deep architectures for {AI}.
\newblock {\em Foundations and Trends in Machine Learning}, 2:1:1--127, 2009.

\bibitem{61}
Yoshua Bengio.
\newblock Practical recommendations for gradient-based training of deep
  architectures.
\newblock In Klaus-Robert~M{\"{u}}ller Gr{\'{e}}goire~Montavon, Genevi{\`{e}}ve
  B.~Orr, editor, {\em Neural Networks: Tricks of the Trade}, pages 437--478.
  Springer, 2012.

\bibitem{60}
Yoshua Bengio, Aaron Courville, and Pascal Vincent.
\newblock Representation learning: A review and new perspectives.
\newblock {\em IEEE Transactions on Pattern Analysis and Machine Intelligence},
  35(8):1798--1828, 2013.

\bibitem{29}
Yoshua Bengio, Pascal Lamblin, Dan Popovici, and Hugo Larochelle.
\newblock Greedy layer-wise training of deep networks.
\newblock {\em Advances in Neural Information Processing Systems}, 19:153--160,
  2007.

\bibitem{23}
Yoshua Bengio and Yann LeCun.
\newblock Scaling learning algorithms towards ai.
\newblock {\em Large-scale kernel machines}, 34:1--41, 2007.

\bibitem{cho2014properties}
Kyunghyun Cho, Bart Van~Merri{\"e}nboer, Dzmitry Bahdanau, and Yoshua Bengio.
\newblock On the properties of neural machine translation: Encoder-decoder
  approaches.
\newblock {\em arXiv preprint arXiv:1409.1259}, 2014.

\bibitem{ChorowskiBSCB15}
Jan Chorowski, Dzmitry Bahdanau, Dmitriy Serdyuk, KyungHyun Cho, and Yoshua
  Bengio.
\newblock Attention-based models for speech recognition.
\newblock {\em CoRR}, abs/1506.07503, 2015.

\bibitem{276}
Farideh Ebrahimi, Mohammad Mikaeili, Edson Estrada, and Homer Nazeran.
\newblock Automatic sleep stage classification based on eeg signals by using
  neural networks and wavelet packet coefficients.
\newblock In {\em Engineering in Medicine and Biology Society, 2008. EMBS 2008.
  30th Annual International Conference of the IEEE}, pages 1151--1154. IEEE,
  2008.

\bibitem{76}
D.~Erhan, Y.~Bengio, A.~Courville, P.A. Manzagol, P.~Vincent, and S.~Bengio.
\newblock Why does unsupervised pre-training help deep learning?
\newblock {\em Journal of Machine Learning Research}, 11:625--660, February
  2010.

\bibitem{258}
Arthur Flexer, Georg Gruber, and Georg Dorffner.
\newblock A reliable probabilistic sleep stager based on a single eeg signal.
\newblock {\em Artificial Intelligence in Medicine}, 33(3):199--207, 2005.

\bibitem{18}
Theo Gasser, Petra B{\"a}cher, and Joachim M{\"o}cks.
\newblock Transformations towards the normal distribution of broad band
  spectral parameters of the eeg.
\newblock {\em Electroencephalography and clinical neurophysiology},
  53(1):119--124, 1982.

\bibitem{167}
A.~L. Goldberger, L.~A.~N. Amaral, L.~Glass, J.~M. Hausdorff, P.~Ch. Ivanov,
  R.~G. Mark, J.~E. Mietus, G.~B. Moody, C.-K. Peng, and H.~E. Stanley.
\newblock {PhysioBank, PhysioToolkit, and PhysioNet}: Components of a new
  research resource for complex physiologic signals.
\newblock {\em Circulation}, 101(23):e215--e220, 2000 (June 13).
\newblock Circulation Electronic Pages:
  http://circ.ahajournals.org/cgi/content/full/101/23/e215.

\bibitem{149}
A.~Graves, A.~Mohamed, and G.~Hinton.
\newblock Speech recognition with deep recurrent neural networks.
\newblock In {\em The 38th International Conference on Acoustics, Speech, and
  Signal Processing (ICASSP)}, 2013.

\bibitem{HermannKGEKSB15}
Karl~Moritz Hermann, Tom{\'{a}}s Kocisk{\'{y}}, Edward Grefenstette, Lasse
  Espeholt, Will Kay, Mustafa Suleyman, and Phil Blunsom.
\newblock Teaching machines to read and comprehend.
\newblock {\em CoRR}, abs/1506.03340, 2015.

\bibitem{4}
Sari-Leena Himanen and Joel Hasan.
\newblock Limitations of {R}echtschaffen and {K}ales.
\newblock {\em Sleep Medicine Reviews}, 4(Issue 2):149--167, April 2000.

\bibitem{8}
Max Hirshkowitz.
\newblock Standing on the shoulders of giants: the {S}tandardized {S}leep
  {M}anual after 30 years.
\newblock {\em Sleep Medicine Reviews}, 4(No. 2):169--179, 2000.

\bibitem{lstm1997}
Sepp Hochreiter and Jürgen Schmidhuber.
\newblock Long short-term memory.
\newblock {\em Neural Computation}, 9(8):1735--1780, 1997.

\bibitem{170}
Eric~J Humphrey, Juan~P Bello, and Yann LeCun.
\newblock Feature learning and deep architectures: new directions for music
  informatics.
\newblock {\em Journal of Intelligent Information Systems}, 41(3):461--481,
  2013.

\bibitem{275}
Suwicha Jirayucharoensak, Setha Pan-Ngum, and Pasin Israsena.
\newblock Eeg-based emotion recognition using deep learning network with
  principal component based covariate shift adaptation.
\newblock {\em The Scientific World Journal}, 2014, 2014.

\bibitem{6}
L.~Johnson, A.~Lubin, P.~Naitoh, C.~Nute, and M.~Austin.
\newblock Spectral analysis of the {EEG} of dominant and non-dominant alpha
  subjects during waking and sleeping.
\newblock {\em Electroencephalography and Clinical Neurophysiology}, 26(Issue
  4):361--370, April 1969.

\bibitem{kalchbrenner2013recurrent}
Nal Kalchbrenner and Phil Blunsom.
\newblock Recurrent continuous translation models.
\newblock In {\em EMNLP}, volume~3, page 413, 2013.

\bibitem{257}
G~Klosh, B~Kemp, Th~Penzel, A~Schlogl, P~Rappelsberger, E~Trenker, G~Gruber,
  J~Zeithofer, B~Saletu, WM~Herrmann, et~al.
\newblock The siesta project polygraphic and clinical database.
\newblock {\em Engineering in Medicine and Biology Magazine, IEEE},
  20(3):51--57, 2001.

\bibitem{270}
Sayaka Kohtoh, Yujiro Taguchi, Naomi Matsumoto, Masashi Wada, Zhi-Li HUANG, and
  Yoshihiro Urade.
\newblock Algorithm for sleep scoring in experimental animals based on fast
  fourier transform power spectrum analysis of the electroencephalogram.
\newblock {\em Sleep and Biological Rhythms}, 6(3):163--171, 2008.

\bibitem{266}
Anna Krakovsk{\'a} and Krist{\'\i}na Mezeiov{\'a}.
\newblock Automatic sleep scoring: A search for an optimal combination of
  measures.
\newblock {\em Artificial Intelligence in Medicine}, 53(1):25 -- 33, 2011.

\bibitem{65}
Martin L\"{a}ngkvist, Lars Karlsson, and Amy Loutfi.
\newblock Sleep stage classification using unsupervised feature learning.
\newblock {\em Advances in Artificial Neural Systems}, 2012, 2012.
\newblock doi:10.1155/2012/107046.

\bibitem{288}
Martin L\"{a}ngkvist, Lars Karlsson, and Amy Loutfi.
\newblock Learning feature representations with a cost-relevant sparse
  autoencoder.
\newblock {\em International Journal of Neural Systems}, 24(8):1 -- 1, 2014.

\bibitem{261}
Martin L\"{a}ngkvist, Lars Karlsson, and Amy Loutfi.
\newblock A review of unsupervised feature learning and deep learning for
  time-series modeling.
\newblock {\em Pattern Recognition Letters}, 42(0):11 -- 24, 2014.

\bibitem{278}
Amy Loutfi, Silvia Coradeschi, Ganesh~Kumar Mani, Prabakaran Shankar, and John
  Bosco~Balaguru Rayappan.
\newblock Electronic noses for food quality: A review.
\newblock {\em Journal of Food Engineering}, 2014.

\bibitem{MnihHGK14}
Volodymyr Mnih, Nicolas Heess, Alex Graves, and Koray Kavukcuoglu.
\newblock Recurrent models of visual attention.
\newblock {\em CoRR}, abs/1406.6247, 2014.

\bibitem{17}
AR~Osborne and A~Provenzale.
\newblock Finite correlation dimension for stochastic systems with power-law
  spectra.
\newblock {\em Physica D}, 35(3):357--81, 1989.

\bibitem{268}
Seral {\"O}z{\c{s}}en.
\newblock Classification of sleep stages using class-dependent sequential
  feature selection and artificial neural network.
\newblock {\em Neural Computing and Applications}, 23(5):1239--1250, 2013.

\bibitem{10}
James Pardey, Stephen Roberts, Lionel Tarassenko, and John Stradling.
\newblock A new approach to the analysis of the human sleep/wakefulness
  continuum.
\newblock {\em Sleep Res.}, 5:201--210, February 1996.

\bibitem{267}
Thomas Penzel, Max Hirshkowitz, John Harsh, Ron~D Chervin, Nic Butkov, Meir
  Kryger, Beth Malow, Michael~V Vitiello, Michael~H Silber, Clete~A Kushida,
  et~al.
\newblock Digital analysis and technical specifications.
\newblock {\em J Clin Sleep Med}, 3(2):109--120, 2007.

\bibitem{33}
E.~Pereda, A.~Gamundi, R.~Rial, and J.~Gonzalez.
\newblock Nonlinear behaviour of human {EEG}: fractal exponent versus
  correlation dimension in awake and sleep stages.
\newblock {\em Neuroscience Letters}, 250(2):91?, 1998.

\bibitem{271}
Huy Phan, Quan Do, The-Luan Do, and Duc-Lung Vu.
\newblock Metric learning for automatic sleep stage classification.
\newblock In {\em Engineering in Medicine and Biology Society (EMBC), 2013 35th
  Annual International Conference of the IEEE}, pages 5025--5028. IEEE, 2013.

\bibitem{7}
A.~Rechtschaffen and A~Kales.
\newblock {\em A Manual of Standardized Terminology, Techniques and Scoring
  System for Sleep Stages of Human Subjects}.
\newblock U.S. Government Printing Office, Washington D.C., 1968.

\bibitem{9}
Claude Robert, Christian Guilpin, and Aym�� Limoge.
\newblock Review of neural network applications in sleep research.
\newblock {\em Journal of neuroscience methods}, 78:187--193, February 1997.

\bibitem{260}
Claude Robert, Christian Guilpin, and Ayme Limoge.
\newblock Review of neural network applications in sleep research.
\newblock {\em Journal of Neuroscience methods}, 79(2):187--193, 1998.

\bibitem{RushCW15}
Alexander~M. Rush, Sumit Chopra, and Jason Weston.
\newblock A neural attention model for abstractive sentence summarization.
\newblock {\em CoRR}, abs/1509.00685, 2015.

\bibitem{32}
N~Schaltenbrand, R~Lengelle, M~Toussaint, R~Luthringer, G~Carelli, A~Jacmin,
  E~Lainey, A~Muzet, and JP~Macher.
\newblock Sleep stage scoring using the neural network model: Comparison
  between visual automatic analysis in normal subjects and patients.
\newblock {\em Sleep}, 19(1):26--35, 1996.

\bibitem{265}
Michael~H Silber, Sonia Ancoli-Israel, Michael~H Bonnet, Sudhansu Chokroverty,
  Madeleine~M Grigg-Damberger, Max Hirshkowitz, Sheldon Kapen, Sharon~A Keenan,
  Meir~H Kryger, Thomas Penzel, et~al.
\newblock The visual scoring of sleep in adults.
\newblock {\em Journal of Clinical Sleep Medicine}, 3(2):121--131, 2007.

\bibitem{279}
Sebastian Stober, Daniel~J. Cameron, and Jessica~A. Grahn.
\newblock Classifying {EEG} recordings of rhythm perception.
\newblock In {\em 15th International Society for Music Information Retrieval
  Conference (ISMIR'14)}, 2014.

\bibitem{269}
Genshiro~A Sunagawa, Hiroyoshi S{\'e}i, Shigeki Shimba, Yoshihiro Urade, and
  Hiroki~R Ueda.
\newblock Faster: an unsupervised fully automated sleep staging method for
  mice.
\newblock {\em Genes to Cells}, 18(6):502--518, 2013.

\bibitem{263}
Krist\'ina {\v S}u{\v s}m\'akov\'a and Anna Krakovsk\'a.
\newblock Discrimination ability of individual measures used in sleep stages
  classification.
\newblock {\em Artificial Intelligence in Medicine}, 44(Issue 3):261--277,
  November 2008.

\bibitem{sutskever2014sequence}
Ilya Sutskever, Oriol Vinyals, and Quoc~V Le.
\newblock Sequence to sequence learning with neural networks.
\newblock In {\em Advances in neural information processing systems}, pages
  3104--3112, 2014.

\bibitem{37}
Graham~W Taylor, Geoffrey~E Hinton, and Sam~T Roweis.
\newblock Modeling human motion using binary latent variables.
\newblock {\em Advances in neural information processing systems}, 19:1345,
  2007.

\bibitem{272}
Dan Wang and Yi~Shang.
\newblock Modeling physiological data with deep belief networks.
\newblock {\em International Journal of Information and Education Technology},
  3, 2013.

\bibitem{274}
Zuoguan Wang, Siwei Lyu, Gerwin Schalk, and Qiang Ji.
\newblock Deep feature learning using target priors with applications in ecog
  signal decoding for bci.
\newblock In {\em Proceedings of the Twenty-Third international joint
  conference on Artificial Intelligence}, pages 1785--1791. AAAI Press, 2013.

\bibitem{35}
D~Wulsin, J~Gupta, R~Mani, J~Blanco, and B~Litt.
\newblock Modeling electroencephalography waveforms with semi-supervised deep
  belief nets: faster classification and anomaly measurement.
\newblock {\em Journal of Neural Engineering}, 8:1741 -- 2552, 2011.

\bibitem{xu2015show}
Kelvin Xu, Jimmy Ba, Ryan Kiros, Kyunghyun Cho, Aaron Courville, Ruslan
  Salakhudinov, Rich Zemel, and Yoshua Bengio.
\newblock Show, attend and tell: Neural image caption generation with visual
  attention.
\newblock In {\em International Conference on Machine Learning}, pages
  2048--2057, 2015.

\bibitem{273}
Wei-Long Zheng, Jia-Yi Zhu, Yong Peng, and Bao-Liang Lu.
\newblock Eeg-based emotion classification using deep belief networks.
\newblock In {\em Multimedia and Expo (ICME), 2014 IEEE International
  Conference on}, pages 1--6, July 2014.

\bibitem{262}
Luk{\'a}{\v{s}} Zoubek, Sylvie Charbonnier, Suzanne Lesecq, Alain Buguet, and
  Florian Chapotot.
\newblock Feature selection for sleep/wake stages classification using data
  driven methods.
\newblock {\em Biomedical Signal Processing and Control}, 2(3):171--179, 2007.

\bibitem{264}
Luk\'a{\v s} Zoubek, Sylvie Charbonnier, Suzanne Lesecq, Alain Buguet, and
  Florian Chapotot.
\newblock A two-steps sleep/wake stages classifier taking into account
  artefacts in the polysomnographic signals.
\newblock In {\em Proc. of the 17th World Congress of The International
  Federation of Automatic Control (IFAC)}, 2008.

\end{thebibliography}

\end{document}